\documentclass[fleqn,usenatbib]{mnras}

\usepackage{newtxtext,newtxmath}

\usepackage[T1]{fontenc}

\DeclareRobustCommand{\VAN}[3]{#2}
\let\VANthebibliography\thebibliography
\def\thebibliography{\DeclareRobustCommand{\VAN}[3]{##3}\VANthebibliography}

\usepackage{graphicx}	
\usepackage{amsmath}	
\usepackage{booktabs,multirow,subcaption,threeparttable}
\usepackage{tikz}
\usetikzlibrary{arrows,positioning}
\usepackage{orcidlink}
\usepackage{xurl}
\usepackage[capitalize]{cleveref}
\usepackage{bold-extra}

\newcommand{\repolink}{\url{https://github.com/yongsukyee/uncertain_blackholemass}} 
\newcommand{\subtext}[2]{\ensuremath{#1_{\text{#2}}}} 
\newcommand{\twodots}{\mathinner {\ldotp \ldotp}} 

\title[Uncertainty Quantification of Virial BH Mass]{Uncertainty Quantification of the Virial Black Hole Mass with Conformal Prediction}

\author[S.~Yong and C.~Ong]{
Suk Yee Yong\,\orcidlink{0000-0002-5204-2902}$^{1,2,3,4,5}$\thanks{E-mail: \href{mailto:sukyee.yong@mq.edu.au}{sukyee.yong@mq.edu.au}}
and Cheng Soon Ong\,\orcidlink{0000-0002-2302-9733}$^{5,6,7}$
\\
$^{1}$ARC Centre of Excellence for All Sky Astrophysics in 3 Dimensions (ASTRO 3D)\\
$^{2}$Astrophysics and Space Technologies Research Centre, Macquarie University, Sydney, NSW 2109, Australia\\
$^{3}$School of Mathematical and Physical Sciences, Macquarie University, Sydney, NSW 2109, Australia\\
$^{4}$Australian Astronomical Optics (AAO), Faculty of Science and Engineering, Macquarie University, Sydney, NSW 2109, Australia\\
$^{5}$Machine Learning and Artificial Intelligence Future Science Platform, CSIRO\\
$^{6}$Data61, CSIRO, Canberra, ACT 2601, Australia\\
$^{7}$School of Computing, The Australian National University, Canberra, ACT 2601, Australia
}

\date{Accepted XXX. Received YYY; in original form ZZZ}

\pubyear{2023}

\begin{document}
\label{firstpage}
\pagerange{\pageref{firstpage}--\pageref{lastpage}}
\maketitle

\begin{abstract}
Precise measurements of the black hole mass are essential to gain insight on the black hole and host galaxy co-evolution.
A direct measure of the black hole mass is often restricted to nearest galaxies and instead, an indirect method using the single-epoch virial black hole mass estimation is used for objects at high redshifts.
However, this method is subjected to biases and uncertainties as it is reliant on the scaling relation from a small sample of local active galactic nuclei.
In this study, we propose the application of conformalised quantile regression (CQR) to quantify the uncertainties of the black hole predictions in a machine learning setting.
We compare CQR with various prediction interval techniques and demonstrated that CQR can provide a more useful prediction interval indicator.
In contrast to baseline approaches for prediction interval estimation, we show that the CQR method provides prediction intervals that adjust to the black hole mass and its related properties.
That is it yields a tighter constraint on the prediction interval (hence more certain) for a larger black hole mass, and accordingly, bright and broad spectral line width source.
Using a combination of neural network model and CQR framework, the recovered virial black hole mass predictions and uncertainties are comparable to those measured from the Sloan Digital Sky Survey.
The code is publicly available at \repolink.
\end{abstract}

\begin{keywords}
black hole physics -- (galaxies:) quasars: general -- (galaxies:) quasars: supermassive black holes -- methods: data analysis -- methods: statistical
\end{keywords}

\section{Introduction}\label{sec:intro}

At the centre of every active galactic nuclei (AGN) is a black hole \citep[e.g.,][]{Kormendy+Richstone:1995,Kormendy+Gebhardt:2001,DiMatteo+:2005}.
The black hole mass, \subtext{M}{BH}, is a crucial quantity in understanding the co-evolution between the black hole and its host galaxy \citep[e.g.,][]{Silk+Rees:1998,Ferrarese+Merritt:2000,Kormendy+Ho:2013}.
However, direct and accurate measurements are very limited to close by galaxies as high spatial resolution is required \citep[e.g.,][]{Kormendy+Gebhardt:2001,Ferrarese+Ford:2005}.

Beyond the local universe, the single-epoch virial mass estimation is applied to estimate the virial black hole mass, \subtext{M}{vir}, which is calibrated empirically using reverberation mapping \citep{Blandford+McKee:1982,Peterson:1993} samples of local AGN \citep[e.g.,][]{Wandel+:1999,Kaspi+:2000,Peterson+:2004}.
This method assumes that the gas in the broad line region (BLR) of the AGN is in Keplerian motion and the virial black hole mass is estimated by
\begin{equation}\label{eq:mbhvir}
\subtext{M}{vir}=\frac{\Delta V^{2} R}{G},
\end{equation}
where $G$ is the gravitational constant and $\Delta V$ is the velocity dispersion of a particular broad emission line often measured by the full width at half maximum (FWHM).
Due to the intensive monitoring at a high cadence over a duration, reverberation mapping for multi-epoch observations are often carried out for a limited number of sources \citep[e.g.,][]{Kaspi+:2000,Bentz+:2009,Bentz+:2013}.
Nonetheless, reverberation mapping studies have also found that there is a relationship between the BLR radius $R$ and monochromatic continuum or line luminosities $L$ \citep{Kaspi+:2000,Kaspi+:2005}, which is used as the basis for single-epoch virial mass estimates \citep[e.g.,][]{McLure+Jarvis:2002,Vestergaard+Peterson:2006,Kollmeier+:2006,Shen+:2008}.
Based on this $R\text{-}L$ relation, the BLR size is derived for a given luminosity and then estimate the \subtext{M}{vir}, in which case \cref{eq:mbhvir} can be rewritten as:
\begin{equation}\label{eq:logmbhvir}
\log\subtext{M}{vir}=a + b\log L + c\log\text{FWHM},
\end{equation}
where $(a,b)$ are the coefficients calibrated from reverberation mapping.
The coefficient $c$ is usually set to 2 based on the virial theorem \citep[but see also][]{Wang+:2009,Marziani+:2013}.
Depending on the redshift of the object, different emission line widths and luminosities are used \citep[e.g.,][]{McLure+Jarvis:2002,McLure+Dunlop:2004,Vestergaard+Peterson:2006,Kollmeier+:2006,Vestergaard+Osmer:2009,Shen+Liu:2012}.
For low redshift sources, this is typically the H$\beta$ and \ion{Mg}{ii} lines, and their respective continuum luminosity measured at rest wavelengths of 5100\,\AA\ and 3000\,\AA.

The majority of reverberation mapping studies have been conducted using H$\beta$ on low redshift AGN \citep[e.g.,][]{Kaspi+:2000,Bentz+:2009,Denney+:2010,Grier+:2012,Bentz+:2013,Barth+:2015,Grier+:2017,Malik+:2023}.
Often for higher redshift, the \ion{Mg}{ii} or \ion{C}{iv} line is utilised.
However, this involves applying additional scaling from the H$\beta$ line to formulate the virial mass based on other lines \citep[e.g.,][]{Shen+:2011}.
There have been efforts to establish the $R\text{-}L$ relation for high redshift AGN \citep[e.g.,][]{Kaspi+:2007,Lira+:2018,Bahk+:2019,Shen+:2019,Hoormann+:2019,Grier+:2019,Homayouni+:2020,Yu+:2022}, though it is still debatable whether the single-epoch \subtext{M}{vir} of these lines are reliable or will need further correction \citep[e.g.,][]{Shen+:2008,Shen+Kelly:2012,Shen+Liu:2012} since they might be affected by non-virial component due to the stratified BLR of the different lines \citep{Murray+:1995,Shen+:2008,Yong+:2016,Yong+:2017}.

There are several limitations and sources of uncertainties in using the single-epoch method that could lead to significant error up to 0.5\,dex in the virial black hole mass \citep[e.g.,][]{Peterson+Bentz:2006,Kelly+Bechtold:2007,Shen+:2008,Shen+Kelly:2010,Shen+Kelly:2012,Shen:2013}.
Some of the common issues are as follows.
First, the relationship between the line width of H$\beta$ and \ion{Mg}{ii} might be non-linear \citep[e.g.,][]{Wang+:2009,Shen+:2011,Marziani+:2013}, which is not accounted for if a constant $c=2$ in \cref{eq:logmbhvir} is applied on \ion{Mg}{ii} line-based \subtext{M}{BH}.
Second, the intrinsic scatter in the $R\text{-}L$ relation calibrated against local reverberation mapped AGN samples using H$\beta$ is $\sim 0.2$\,dex \citep{Bentz+:2013} and can be larger than 0.36\,dex when using \ion{Mg}{ii} \citep{Homayouni+:2020}.
The \subtext{M}{vir} based on \ion{Mg}{ii} line have to be properly calibrated such that they match those of H$\beta$ line \citep{McGill+:2008,Wang+:2009,Woo+:2018,Bahk+:2019}.
Various prescriptions have been proposed \citep[e.g.,][]{McLure+Jarvis:2002,McLure+Dunlop:2004,Greene+Ho:2005,Vestergaard+Peterson:2006,Kollmeier+:2006,Vestergaard+Osmer:2009,Shen+:2011} to calibrate the $(a,b)$ coefficients in \cref{eq:logmbhvir}, which can vary depending on which specific line is used \citep{McGill+:2008}.
Practically, this also assumes that a single best fit line from the empirical relationship, fixed by some constant coefficients, is applicable to every sources.
Third, the derived continuum and spectral line properties rely on the choice of spectral fitting process.
Mainly, this requires a consistent procedure for fitting the continuum and modelling individual spectral line component as this will substantially affect the line measurements \citep[e.g.,][]{Shen+:2008}.
The presence of strong absorption lines and using low quality signal-to-noise ratio spectra are likely to result in unreliable measurements \citep{Denney+:2009}.

Recently, several studies have employed machine learning and deep learning methods to predict the properties of the black hole.
\citet{He+:2022} explored the \subtext{M}{BH} correlation with their host galaxy properties using Lasso regression.
They then used the extracted subset of properties to derive an empirical formula for the black hole mass and shown that it is able to retrieve the masses with a scatter of 0.5\,dex.
Though only trained using a small sample available from reverberation mapping, \citet{Eilers+:2022} demonstrated that they are able to generate quasar spectra along with the associated physical properties even for missing spectral region and without requiring calibration from the $R\text{-}L$ scaling relation.
They applied a multi-output Gaussian process latent variable model and estimated the uncertainties in the predicted \subtext{M}{BH} due to errors from measurements and input spectra, and reported a scatter of 0.4\,dex in the predictions.
Similarly, \citet{Chainakun+:2022} used a multi-layer perceptron regressor on a few reverberation mapped AGN samples probed in the X-ray regime and recovered the \subtext{M}{BH} within $\pm (2\text{--}5)\%$.
\citet{Lin+:2023} employed a hybrid deep neural network model consisting of convolutional and fully connected layers on quasar light curves as an alternative to the expensive spectral data.
They predicted the \subtext{M}{BH} from the light curves within 0.37\,dex scatter.

Previous studies primarily considered recovering the black hole mass from the measured \subtext{M}{BH} using light curves or calibrated based on reverberation mapping of low redshift quasars.
However, a question still remain: since all measurements of the black hole mass have intrinsic scatter, \textit{how good is then the uncertainties of the black hole mass predictions?} In this work, we do not attempt to build a more accurate predictor for the black hole mass.
Instead, we focus on quantifying the uncertainties of the line-based virial mass, \subtext{M}{vir}, and address some of the aforementioned limitations and sources of uncertainty.
In particular, we employ a conformal prediction for regression framework, specifically the conformalised quantile regression \citep{Romano+:2019}, and conduct a comparative study with several other prediction interval approaches.
The conformalised quantile regression is of particular interest as it has been shown to be flexible to any heteroscedasticity in the data and generates adaptive prediction intervals.

We present a method to quantify the uncertainty in black hole mass predictions with adaptive prediction intervals.
We separate this into two parts:
\begin{enumerate}
  \item Perform representation learning (finding a good feature encoding) using a neural network model: This effectively avoid the need to fit and obtain individual line measurements.
  \item Generate predictions and prediction intervals for the line-based \subtext{M}{vir}: We examine different prediction interval methods to quantify the uncertainties in the \subtext{M}{vir}.
\end{enumerate}

The outline of the paper is as follows.
\Cref{sec:dataset} describes the dataset utilised.
Overviews of the neural network model and the prediction interval methods employed are given in \cref{sec:mvirppi}.
The results followed by discussions in \cref{sec:results} and \cref{sec:discussions}, respectively.
Finally, \cref{sec:summary} summarises our findings.

\section{Dataset}\label{sec:dataset}

We briefly describe the dataset used in this work and pre-processing applied on the data.
We use the recent catalogue of quasar properties \citep{Wu+Shen:2022} derived from the Sloan Digital Sky Survey (SDSS) Data Release 16 Quasar \citep[DR16Q;][]{Lyke+:2020} catalogue.
The data\footnote{\url{http://quasar.astro.illinois.edu/paper_data/DR16Q/}} and tutorial\footnote{\url{https://github.com/QiaoyaWu/sdss4_dr16q_tutorial}} containing the description of the data and demo are publicly available online.

The details on the derived spectral line measurements are described in Section~3 of \citet{Wu+Shen:2022} and also in their earlier work \citep[e.g.,][]{Shen+:2019}, which we briefly outline here.
They corrected the spectra for Galactic reddening using the dust map from \citet{Schlegel+:1998} and \citet{Schlafly+:2011} and the extinction curve from \citet{Cardelli+:1989}.
After shifting the spectra to the rest-frame using the redshift from the SDSS DR16Q catalogue, they fitted the continuum by a power law and a third-order polynomial, and also an iron template \citep{Boroson+Green:1992,Vestergaard+Wilkes:2001,Tsuzuki+:2006,Salviander+:2007} to several continuum fitting windows that are not affected by broad line emission.
Quasars that have peculiar continuum shapes are fitted with the additive (positive-definite) polynomial component.
They subtracted the continuum and iron fit from the spectrum to form a line-only spectrum, which is then fitted with a set of Gaussians in logarithmic wavelength space.
To minimise the effect of absorption lines from intervening absorption systems, they performed an iterative approach to mask pixels below 3-sigma of the original model fit and refit.

From the best spectral fitting parameters, \citet{Wu+Shen:2022} measured the continuum and emission line properties, including the spectral line peak and FWHM.
Using a Monte Carlo approach, they estimated the uncertainties in the line measurements by randomly perturb the original spectrum at each pixel with a Gaussian.
They performed for 25 iterations and took the semi-amplitude within the 16th and 84th percentile range as the error of each spectral quantity.
To calibrate the coefficient $(a,b)$ for the single-epoch \subtext{M}{vir}, they adopted $(0.91, 0.50)$ for H$\beta$ \citep{Vestergaard+Peterson:2006} and $(0.74, 0.62)$ for \ion{Mg}{ii} \citep{Shen+:2011}.
The measurement uncertainties in the \subtext{M}{vir} are also provided in the catalogue.
Their compiled catalogue has a total of 750,414 spectra, with each data file containing the original fluxes, continuum fluxes, and the spectral line fluxes with continuum subtracted.

\begin{figure}
\centering
\includegraphics[width=\linewidth]{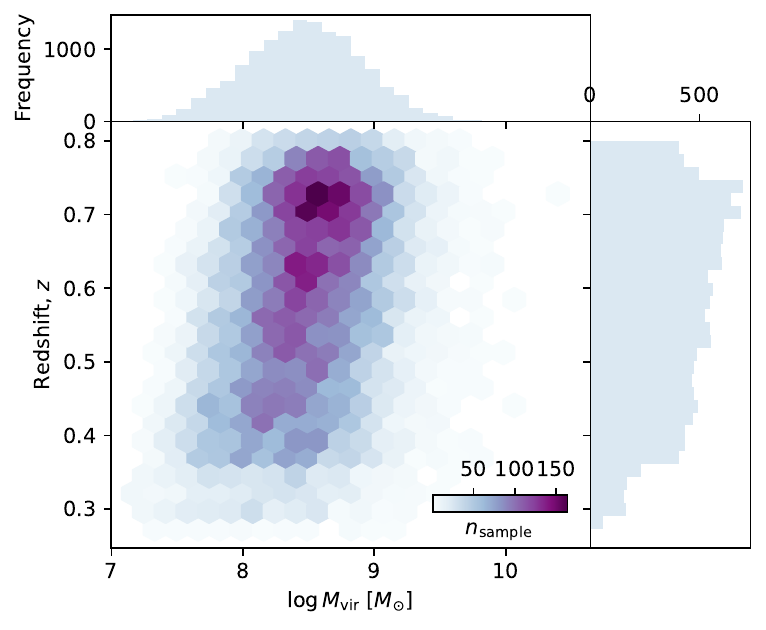}
\caption{Sample distribution of virial black hole masses with redshifts.
The density of the sample indicated by the number of samples, \subtext{n}{sample}, is shown in coloured with darker gradient being more dense.}
\label{fig:datamvirvsz}
\end{figure}

For the sample selection, we follow the recommended quality cuts for specific emission lines in their paper, namely line flux/flux error $>2$ and logarithm line luminosity ranges 38--48\,erg\,s$^{-1}$ and apply them to the H$\beta$ and \ion{Mg}{ii} lines.
We further restrain the sample that have both H$\beta$ and \ion{Mg}{ii} line widths and black hole masses available.
As black hole mass is derived from the line width, we remove quasars with large errors in the black hole mass with $>0.5$\,dex and line width error with $>2000\,$km\,s$^{-1}$.
We also select spectra with high median signal-to-noise ratio per pixel of $\geq 10$.
A summary of selection criteria along with the number of drop out after each cut is listed in \cref{apd:sampleselect}.
Our final data sample consists of 13,952 spectra, and the distributions of the black hole masses with redshifts are shown in \cref{fig:datamvirvsz}.

We split the data into 70\% training, 20\% validation, and 10\% test sets, which are 9766, 2930, and 1256 spectra, respectively.
In training the machine learning model, we find that using the fluxes of the entire spectrum as our input data does not lead to any meaningful feature extraction.
This might be due to the noisy fluctuations and spurious spectral spikes in the fluxes.
Hence, we use the spectral line flux with continuum subtracted, which is provided in the data file from \citet{Wu+Shen:2022}, as the input for the training and validation of the machine learning.
The validation set is used for evaluating the model performance during the training.
Since we do not utilise the wavelength, which contains the position information, when training the neural network and the line flux mainly cut off at $\sim 1000$ pixels, we therefore truncate the data to the first 1000 pixels.
The virial H$\beta$ and \ion{Mg}{ii} black hole mass estimates are used as the ground truth labels.
The fluxes and labels are normalised from 0 to 1.

\section{Virial Black Hole Mass Predictions and Uncertainties}\label{sec:mvirppi}

\begin{figure*}
\centering
\includegraphics[width=\linewidth]{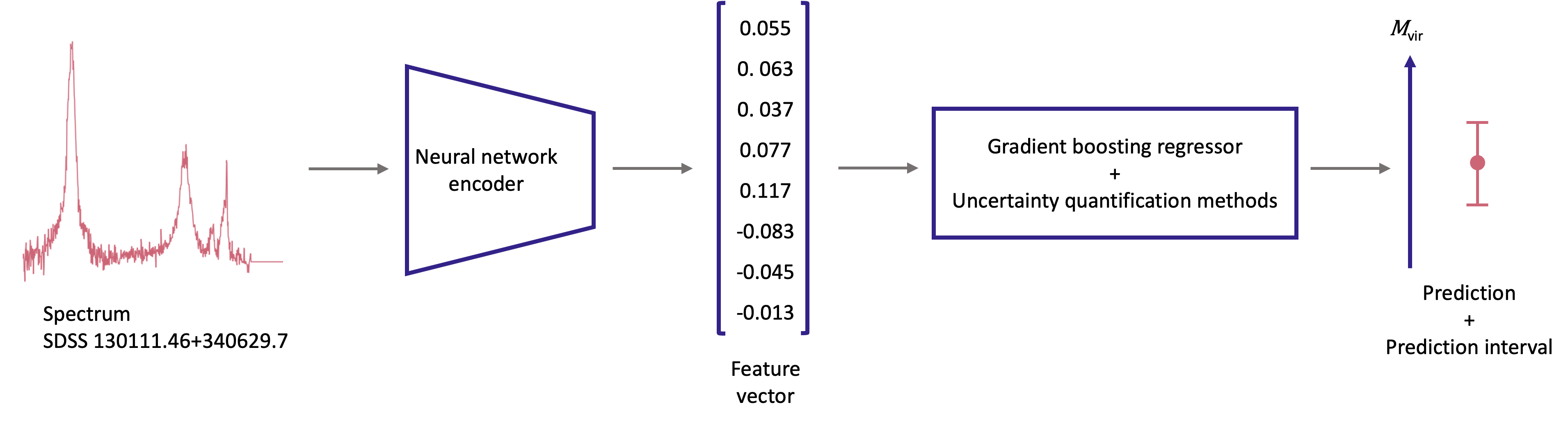}
\caption{A flowchart of the end-to-end prediction pipeline.
First, the input spectra of dimension 1000 are fed into the fully connected neural network encoder.
The encoder consists of 2 hidden layers of 64 neurons each with rectified linear unit as activation function and dropout of probability 0.1.
Another layer of 8 neurons is added for feature extraction, which outputs a vector of 8 features for each spectrum.
Gradient boosting for regression and uncertainty quantification methods are performed to yield predictions and prediction intervals for the virial black hole mass, \subtext{M}{vir}.}
\label{fig:pipelineflowchart}
\end{figure*}

In this section, we detail the end-to-end pipeline implemented from training the input data using a neural network model to output the prediction intervals.
A flowchart of the pipeline is illustrated in \cref{fig:pipelineflowchart}.

The following notation is adopted.
Given $n$, the number of independently and identically distributed training data with input-target pair $\{(X_{i},Y_{i})\}_{i=1}^{n}$, we perform regression.
In regression analysis, the target can be represented by
\begin{equation}
Y=\hat{\mu}(X)+\epsilon,
\end{equation}
where $\hat{\mu}(X)$ is the regression function to be estimated and $\epsilon$ is the model error.
In this case, the target $Y$ is the virial black hole mass \subtext{M}{vir}, and the input $X$ is the SDSS spectra.

\subsection{Construction of neural network for feature extraction}

To extract the feature vectors from the spectra, we employ a supervised learning approach using a generic fully connected neural network model with similar hidden layer architecture in \citet{Romano+:2019}.
The neural network is constructed using \textsc{PyTorch} \citep{pytorch:2019}, an open source machine learning framework in \textsc{Python}.
The input layer consists of 1000 neurons followed by 3 hidden layers of 64, 64, and 8 neurons with rectified linear unit as activation function and dropout \citep{Srivastava+:2014} of probability 0.1, then finally an output layer with 1 node and sigmoid activation function.
The outputs of the second to last layer of 8 neurons is saved as features of the spectra.
As our main aim is not to find the best model, we do not attempt any optimisation or hyperparameters tuning on the model.
Following \citet{Romano+:2019}, the stochastic gradient descent based Adam optimiser \citep{Kingma+Ba:2015} is used with initial learning rate of $5 \times 10^{-4}$ and weight decay regularisation parameter of $10^{-6}$.
Additionally, we apply a constant learning rate scheduler that decreases by a factor of 0.5 every 2 steps.
The model is optimised with mean squared error (MSE) as the cost function.
The model is then trained for 100 epochs with batch size of 64.

There are a few main assumptions that we made in training our machine learning model.
We assume that the SDSS spectra are of good quality with reliable derived properties.
On the other hand, note that these properties are also constrained by the same assumptions used to derive them.
In particular, the derived \subtext{M}{vir} from SDSS are dependent on various factors, including the Keplerian motion assumption into the virial theorem and the applicability of the empirical scaling relation to single-epoch mass estimates.
There are also potential systematic uncertainties that might not be completely accounted for.
Further caveats are discussed in \cref{ssec:furtherexpcaveats}.
To train the supervised neural network model, we use the spectra as inputs and the SDSS DR16Q derived virial H$\beta$ and \ion{Mg}{ii} black hole mass estimates as targets to be optimised.
The uncertainties of the measured \subtext{M}{vir} are not included when training the model.

\subsection{Construction of regressor for predictions}

After the feature extraction process from the neural network, we use gradient boosting for regression to make the predictions.
Depending on the uncertainty quantification methods, which will be described in \cref{ssec:constructpi}, the quantile loss is applied for the conformalised quantile regression, while the MSE loss for the rest of the resampling techniques.
To reduce the prediction error, we optimise the regressor by performing a randomised search with 10-fold cross-validation for 100 iterations to find the best hyperparameters for the regressor.
The explored parameter space and the adopted best model are shown in \cref{tab:regressorhyperparam}.

\begin{table*}
  \centering
  \caption{Explored parameter space for the gradient boosting regressor and the best hyperparameter values found using randomised search with 10-fold cross-validation.
  The regressor for conformalised quantile regressor (CQR) uses quantile loss, while other uncertainty quantification methods use mean squared error (MSE) loss.}
  \label{tab:regressorhyperparam}
  \begin{tabular}{lccccc}
  \toprule
  Hyperparameter & Search Range & \multicolumn{2}{c}{Regressor for H$\beta$-based Data} & \multicolumn{2}{c}{Regressor for \ion{Mg}{ii}-based Data} \\
  \cmidrule(lr){3-4} \cmidrule(lr){5-6}
   & & MSE loss & Quantile loss & MSE loss & Quantile loss \\
  \midrule
  \texttt{learning\_rate} & \texttt{Uniform(0,1)} & 0.013 & 0.051 & 0.013 & 0.051 \\
  \texttt{max\_depth} & $2 \twodots 30$ & 26 & 20 & 26 & 20 \\
  \texttt{max\_leaf\_node} & $2 \twodots 50$ & 15 & 24 & 15 & 24 \\
  \texttt{n\_estimators} & $10 \twodots 500$ & 251 & 152 & 251 & 152 \\
  \bottomrule
  \end{tabular}
\end{table*}

To check the performance of the regression model, two common evaluation metrics, the mean absolute error (MAE) and root mean squared error (RMSE), are evaluated.
MAE is the average of the absolute errors between the target value $Y_{i}$ and predicted value $\hat{\mu}(X_{i})$:
\begin{equation}\label{eq:mae}
\mathrm{MAE}=\frac{1}{n}\sum_{i=1}^{n} |Y_{i}-\hat{\mu}(X_{i})|.
\end{equation}
RMSE is the average of the squares of the difference between the target and predicted value:
\begin{equation}\label{eq:rmse}
\mathrm{RMSE}=\sqrt{\frac{1}{n}\sum_{i=1}^{n}\left[Y_{i}-\hat{\mu}(X_{i})\right]^{2}}.
\end{equation}
MAE is more robust to outliers, while the RMSE is easier to optimise.
In both cases, the lower score the better.
Additionally, 10-fold cross validation is performed to obtain the mean and standard deviation of the respective evaluation metrics.

\subsection{Assessing the performance of prediction intervals}\label{ssec:performance}

The two criteria that are crucial to assess the performance of the prediction intervals are the coverage and the width \citep[e.g.,][]{Khosravi+:2010}.
The prediction interval coverage probability (PICP) or coverage for short reflects the probability that the prediction interval will contain the target value, which is defined as
\begin{equation}
\mathrm{PICP}=\frac{1}{\subtext{n}{test}}\sum_{i=1}^{\subtext{n}{test}}c_{i},
\end{equation}
where $c_{i}=1$ if $Y_{i} \in [L(X_{i}), U(X_{i})]$ otherwise $c_{i}=0$, $L(X_{i})$ and $U(X_{i})$ are the lower and upper bounds, respectively.
Ideally, higher PICP is better and it should be close to the nominal confidence level of $(1-\alpha)$.
The confidence level is set to be 90\%.
Additionally, we compute the coefficient of determination, $R^{2}$, to measure the percentage of variance 
between PICP and the $(1-\alpha)$ nominal coverage rate.
\begin{equation}
R^{2}=1-\frac{\sum\left[Y_{i}-\hat{\mu}(X_{i})\right]^{2}}{\sum(Y_{i}-\bar{Y})^{2}},
\end{equation}
where $\bar{Y}$ is the mean of $Y$.
The $R^{2}$ ranges 0--1 (or in percentage 0--100\%), where the higher the better with 100\% being a perfect fit.

The mean prediction interval width (MPIW) measures the wideness of the prediction interval and is given by the average of the width
\begin{equation}
\mathrm{MPIW}=\sum_{i=1}^{\subtext{n}{test}} \left[U(X_{i})-L(X_{i})\right],
\end{equation}
where the prediction interval width is defined as the difference between the upper and lower bounds, which is the term in the square bracket.
The larger the width, the more uncertain.
It is desirable to have a high PICP but a narrow MPIW.

\subsection{Construction of prediction intervals}\label{ssec:constructpi}

Various methods to construct prediction intervals have been developed and a comparison between different strategies is reviewed in \citet{FoygelBarber+:2021}.
Using an open-source \textsc{Python} package called model agnostic prediction interval estimator \citep{Taquet+:2022} or \textsc{MAPIE}\footnote{\url{https://github.com/scikit-learn-contrib/MAPIE}}, we explore different techniques to estimate the prediction intervals.

To estimate the prediction interval, the model error $\epsilon$ can be characterised as the conditional probability distribution of $Y$ given $X$, $\mathbb{P}_{Y|X}$.
In practice, this is estimated by the difference between the label and the prediction, $Y_{i} - \hat{\mu}(X_{i})$.
Let $(X_{n+1},Y_{n+1})$ be the input-target for a new unseen test point.
Suppose we want to construct a valid prediction interval $\hat{\mathcal{C}}_{n,\alpha}(X_{n+1})$ for the test data.
It should satisfy
\begin{equation}\label{eq:coverageguarantee}
\mathbb{P}\{Y_{n+1} \in \hat{\mathcal{C}}(X_{n+1})\} \geq 1-\alpha,
\end{equation}
where $\alpha$ is the target quantile and the complementary $(1-\alpha)$ is the confidence level or coverage rate.
The estimator (for the prediction interval) is considered calibrated if it satisfies the inequality in \cref{eq:coverageguarantee}.
A conformity score is a measure of how similar a sample is compared to the rest of the dataset and is used to determine the threshold for the quantile, leading to a prediction interval.

A key challenge to estimating the prediction interval is to ensure statistical consistency, and various approaches have been proposed.
Conformal prediction \citep{Vovk+:1999,Papadopoulos+:2002,Vovk+:2005,Lei+Wasserman:2014,Angelopoulos+Bates:2021} offers a robust uncertainty quantification framework and a distribution-free coverage guarantee that satisfy \cref{eq:coverageguarantee}.
As a set of baseline comparison, we compare conformal prediction against various uncertainty quantification methods from the \textsc{MAPIE} package, namely naive, jackknife+-after-bootstrap, cross-validation and its variations.
We briefly review the methods we use in this paper in the following.

\subsubsection{``Naive'' conformity score}

Consider a simple or ``naive'' way to compute conformity score, by using the residual of the training dataset, which gives
\begin{equation}
\hat{\mathcal{C}}_{n,\alpha}^{\mathrm{naive}}(X_{n+1})=\hat{\mu}(X_{n+1}) \pm \hat{q}_{n,\alpha}^{+}|Y_{i}-\hat{\mu}(X_{i})|,
\end{equation}
where $\hat{q}_{n,\alpha}^{+}$ is the $(1-\alpha)$ quantile of the empirical distribution.
Though this method is computationally cheap, it does not guarantee coverage and is likely to overfit, which underestimates the prediction interval widths.

\subsubsection{Jackknife+-after-bootstrap}

The standard jackknife is a leave-one-out cross-validation \citep[CV;][]{Quenouille:1949,Quenouille:1956,Tukey:1958,Miller:1974,Stone:1974} approach.
We opt for jackknife+-after-bootstrap \citep[jackknife+ab;][]{Kim+:2020} as it is more computationally efficient than the standard jackknife.
The steps to infer the jackknife+ab prediction intervals are as follow:
\begin{itemize}
  \item Bootstrap resampling from the training set with replacement $K$ times, $B_{1},\dots,B_{K}$.
  \item Fit the $K$ regression functions $\hat{\mu}_{B_{k}}$ on the bootstrapped dataset.
  \item Aggregate the estimated prediction function using the bootstrapped dataset excluding sample $i$ given by $\hat{\mu}_{\varphi,-i}=\varphi(\{\hat{\mu}_{B_{k}}(X_{n+1}: i \notin B_{k})\})$, where $\varphi$ is the aggregation function usually taken to be the mean or median.
  The mean is used, which is the default.
  Then compute the conformity score as the residual $R_{\varphi,i}=|Y_{i}-\hat{\mu}_{\varphi,-i}(X_{i})|$ for $i=1,\dots,n$.
  \item Output jackknife+ab prediction interval:
  \begin{equation}
  \begin{split}
  \hat{\mathcal{C}}_{n,\alpha,B}^{\mathrm{jackknife+ab}}(X_{n+1})=\left[\hat{q}_{n,\alpha}^{-}\{\hat{\mu}_{\varphi,-i}(X_{n+1}) - R_{\varphi,i}\}, \right. \\
  \left. \hat{q}_{n,\alpha}^{+}\{\hat{\mu}_{\varphi,-i}(X_{n+1}) + R_{\varphi,i}\}\right],
  \end{split}
  \end{equation}
  where $\hat{q}_{n,\alpha}^{-}$ is the $\alpha$ quantile of the distribution and recall the $(1-\alpha)$ counterpart is $\hat{q}_{n,\alpha}^{+}$.
\end{itemize}

\subsubsection{Cross-validation and its variations}

Rather than the leave-one-out method, cross validation can be performed in $K$-fold to reduce computation time.
The steps to infer the CV+ prediction intervals are as follow:
\begin{itemize}
  \item Split training set into $K$ disjoint subsets $S_{1},\dots,S_{K}$ each of size $m=n/K$.
  \item Fit the $K$ regression functions $\hat{\mu}_{-S_{k}}$ on the training dataset with $k$th subset excluded.
  \item Compute the conformity score from the $K$-fold process as $R_{i}^{\mathrm{CV}}=|Y_{i}-\hat{\mu}_{-S_{k(i)}}(X_{i})|$, where the subset $k(i)$ contains $i$.
  \item Output CV+ prediction interval:
  \begin{equation}
  \begin{split}
  \hat{\mathcal{C}}_{n,\alpha,K}^{\mathrm{CV+}}(X_{n+1})=\left[\hat{q}_{n,\alpha}^{-}\{\hat{\mu}_{-S_{k(i)}}(X_{n+1}) - R_{i}^{\mathrm{CV}}\}, \right. \\
  \left. \hat{q}_{n,\alpha}^{+}\{\hat{\mu}_{-S_{k(i)}}(X_{n+1}) + R_{i}^{\mathrm{CV}}\}\right].
  \end{split}
  \end{equation}
\end{itemize}
The jackknife+ab and CV+ provide slightly larger coverage guarantee of $(1-2\alpha)$.

For standard CV, the output prediction interval is defined as
\begin{equation}
\hat{\mathcal{C}}_{n,\alpha}^{\mathrm{CV}}(X_{n+1})=\left[\hat{q}_{n,\alpha}^{-}\{\hat{\mu}(X_{n+1}) - R_{i}^{\mathrm{CV}}\}, \hat{q}_{n,\alpha}^{+}\{\hat{\mu}(X_{n+1}) + R_{i}^{\mathrm{CV}}\}\right].
\end{equation}

Another variation of CV that is more conservative than CV+ is the CV-minmax method given by
\begin{equation}
\begin{split}
\hat{\mathcal{C}}_{n,\alpha}^{\mathrm{CV-minmax}}(X_{n+1})=\left[\min_{i=1,\dots,n} \hat{\mu}_{-i}(X_{n+1}) - \hat{q}_{n,\alpha}^{+}\{R_{i}^{\mathrm{CV}}\}, \right. \\
\left. \max_{i=1,\dots,n} \hat{\mu}_{-i}(X_{n+1}) + \hat{q}_{n,\alpha}^{+}\{R_{i}^{\mathrm{CV}}\}\right],
\end{split}
\end{equation}
which guarantee the $(1-\alpha)$ coverage in \cref{eq:coverageguarantee}

\subsubsection{Conformalised quantile regression}

As the transductive or full conformal prediction is computationally heavy, the inductive or split conformal prediction \citep{Papadopoulos+:2002,Papadopoulos:2008} approach is applied to alleviate the issue.
In this setting, it trains the model only once, but requires data splitting for the calibration set.
For regression, the conformalised quantile regression \citep[CQR;][]{Romano+:2019} is built upon conformal prediction and quantile regression \citep{Koenker+Bassett:1978} to provide a two-sided prediction interval or band.
The steps to infer the CQR prediction intervals are as follow:
\begin{itemize}
  \item Split dataset into two disjoint subsets for training set $\mathcal{I}_{1}$ and calibration set $\mathcal{I}_{2}$.
  \item Fit two conditional quantile functions for the lower quantile $\hat{q}_{\alpha/2}$ and upper quantile $\hat{q}_{1-\alpha/2}$.
  \item Compute the conformity score for each $i \in \mathcal{I}_{2}$ as $E_{i}^{\mathrm{CQR}}=\max\{\hat{q}_{\alpha/2}(X_{i})-Y_{i}, Y_{i}-\hat{q}_{1-\alpha/2}(X_{i})\}$.
  \item Compute $\hat{Q}_{1-\alpha}(E^{\mathrm{CQR}},\mathcal{I}_{2}):=(1-\alpha)(1+1/|\mathcal{I}_{2}|)$-th empirical quantile of $\{E_{i}^{\mathrm{CQR}}: i \in \mathcal{I}_{2}\}$.
  \item Output CQR prediction interval:
  \begin{equation}
  \begin{split}
  \hat{\mathcal{C}}_{n,\alpha}^{\mathrm{CQR}}(X_{n+1})=\left[\hat{q}_{\alpha/2}(X_{n+1}) - \hat{Q}_{1-\alpha}(E^{\mathrm{CQR}},\mathcal{I}_{2}), \right. \\
  \left. \hat{q}_{1-\alpha/2}(X_{n+1}) + \hat{Q}_{1-\alpha}(E^{\mathrm{CQR}},\mathcal{I}_{2})\right].
  \end{split}
  \end{equation}
\end{itemize}
We employ CQR with inductive split using the validation set as the calibration set.

Towards the final stage of the prediction pipeline in \cref{fig:pipelineflowchart}, prediction intervals are obtained from the various uncertainty quantification methods.
Their performances are evaluated and compared using the two metrics, PICP and MPIW, as defined previously in \cref{ssec:performance}.

\section{Results}\label{sec:results}
\subsection{Effectiveness of neural network for feature extraction}

\begin{figure*}
\centering
\begin{subfigure}[t]{0.43\linewidth}
  \includegraphics[width=\linewidth]{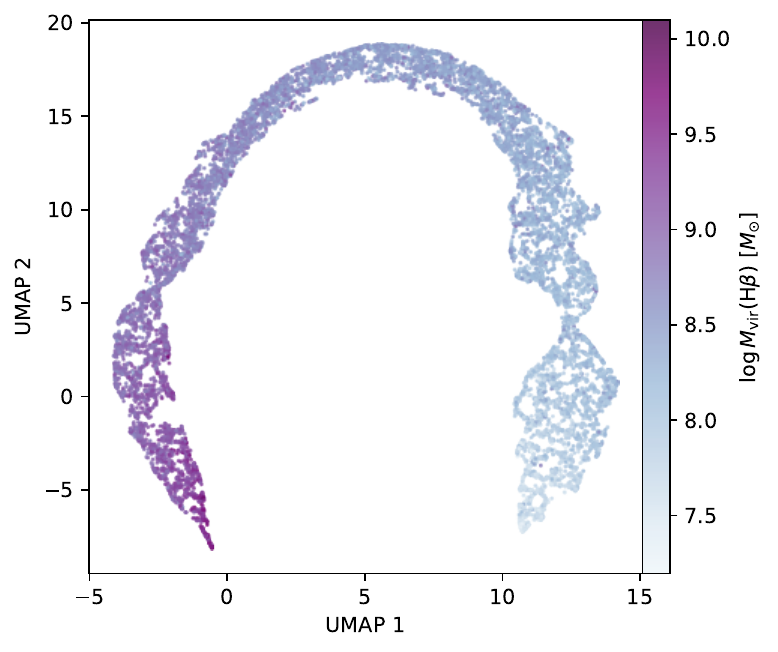}
  \caption{For H$\beta$}
  \label{fig:umap2d_hbeta}
\end{subfigure}
\qquad
\begin{subfigure}[t]{0.43\linewidth}
  \includegraphics[width=\linewidth]{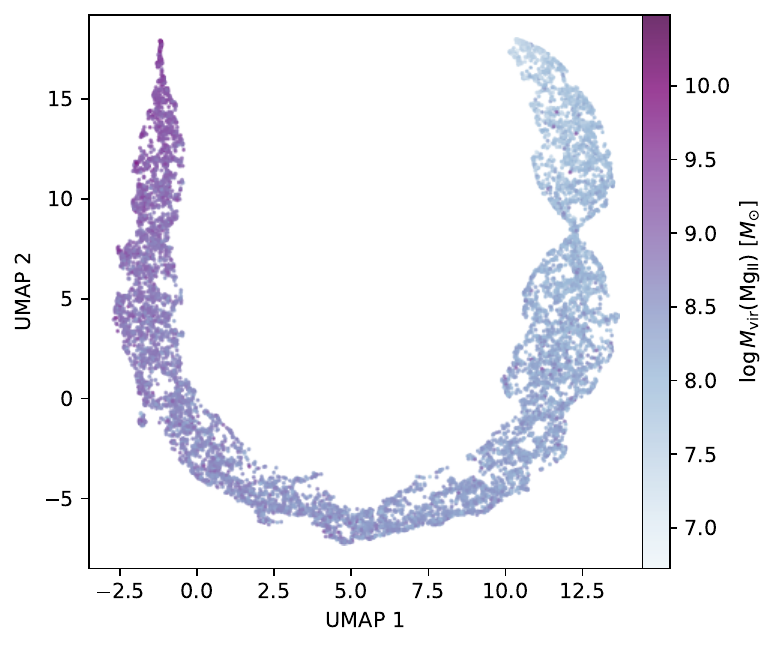}
  \caption{For \ion{Mg}{ii}}
  \label{fig:umap2d_mgii}
\end{subfigure}
\caption{Uniform Manifold Approximation and Projection (UMAP) representation of the parameter space in 2-dimension from the 8-dimension feature extraction of the training dataset, color-coded by actual \subtext{M}{vir}.}
\label{fig:umap2d}
\end{figure*}

Features extracted by a neural network are not directly interpretable, as they do not correspond to any particular physical parameters.
However, if the regressor is to perform well, the extracted features should capture meaningful aspects of the raw data.
To determine whether the extracted features from the neural network are meaningful, we use Uniform Manifold Approximation and Projection \citep{McInnes+:2018} or UMAP\footnote{\url{https://github.com/lmcinnes/umap}}, a dimension reduction technique to project the 8-dimension features to 2-dimension parameter space.
As the purpose is purely for visualisation, we set the number of components to 2 and use the defaults for the rest of the UMAP hyperparameters.
It can be observed in \cref{fig:umap2d} that the 2-dimensional UMAP representation is structured such that smaller \subtext{M}{vir} objects tend to be on the right and gradually towards the left for increasing \subtext{M}{vir}.
This affirms that the 8 features extracted are sensible to characterise the H$\beta$ and \ion{Mg}{ii} line-based \subtext{M}{vir}, which are used as inputs for the regressor.

\subsection{Performance of regressor for predictions}

Due to the various assumptions imposed on estimating the black hole mass, the black hole mass estimates can be substantially biased and uncertain \citep[e.g.,][]{Kelly+Bechtold:2007,Shen+:2008,Shen+Kelly:2010}.
To leverage the need for individual spectral line fitting, we use a neural network model to extract the latent feature vectors and use a regressor to predict the \subtext{M}{vir}.
The prediction errors of the regressors are shown in \cref{tab:prederror}.
Overall, the performances of the regressors using quantile loss and MSE loss for both H$\beta$ and \ion{Mg}{ii} cases are similar.
The \ion{Mg}{ii} line-based \subtext{M}{vir} prediction errors are slightly larger compared to those of H$\beta$.
As previously mentioned, this is likely because the \subtext{M}{vir} based on H$\beta$ is better calibrated \citep[e.g.,][]{Shen+:2011}, which leads to the smaller prediction error.

As a comparison, the performances of the black hole mass predictions trained using machine learning model reported by other studies are also listed in \cref{tab:prederror}.
It can be seen that the predictions are relatively good with low prediction errors when compared to those from other studies.
Though note that the dataset used in those studies are not the same from one another; thus, the difference in scores might also be attributed to the difficulty of the machine learning task.

\begin{table*}
  \centering
  \caption{Evaluation metrics for the H$\beta$ and \ion{Mg}{ii} line-based virial black hole mass predictions.
  The gradient boosting regressor is used as the regressor with quantile loss for CQR, while mean squared error (MSE) loss for other uncertainty quantification methods.
  The scores are the means and standard deviations evaluated by 10-fold cross-validation.
  Lower score is better.
  For comparison, the evaluation scores of the predicted black hole mass from other studies are provided in the last column.}
  \label{tab:prederror}
  \begin{tabular}{l*{5}{c}}
  \toprule
  Scoring Metric & \multicolumn{2}{c}{Regressor for H$\beta$-based Data} & \multicolumn{2}{c}{Regressor for \ion{Mg}{ii}-based Data} & Other Studies$^{\dag}$ \\
  \cmidrule(lr){2-3} \cmidrule(lr){4-5}
   & MSE loss & Quantile loss & MSE loss & Quantile loss \\
  \midrule
  MAE & $0.144 \pm 0.013$ & $0.144 \pm 0.012$ & $0.169 \pm 0.011$ & $0.169 \pm 0.011$ & 0.010--0.260 [\citenum{Chainakun+:2022}] \\
  RMSE & $0.198 \pm 0.026$ & $0.198 \pm 0.024$ & $0.222 \pm 0.017$ & $0.222 \pm 0.014$ & 0.500 [\citenum{He+:2022}], 0.373 [\citenum{Lin+:2023}] \\
  \bottomrule
  \multicolumn{6}{l}{$^{\dag}$ Note the different dataset and machine learning model used in these studies}\\
  \multicolumn{6}{l}{[\citenum{Chainakun+:2022}] \citet{Chainakun+:2022}, [\citenum{He+:2022}] \citet{He+:2022}, [\citenum{Lin+:2023}] \citet{Lin+:2023}}\\
  \end{tabular}
\end{table*}

\subsection{Reliability of prediction intervals}

\begin{figure*}
\centering
\begin{subfigure}[t]{0.83\linewidth}
  \includegraphics[width=\linewidth]{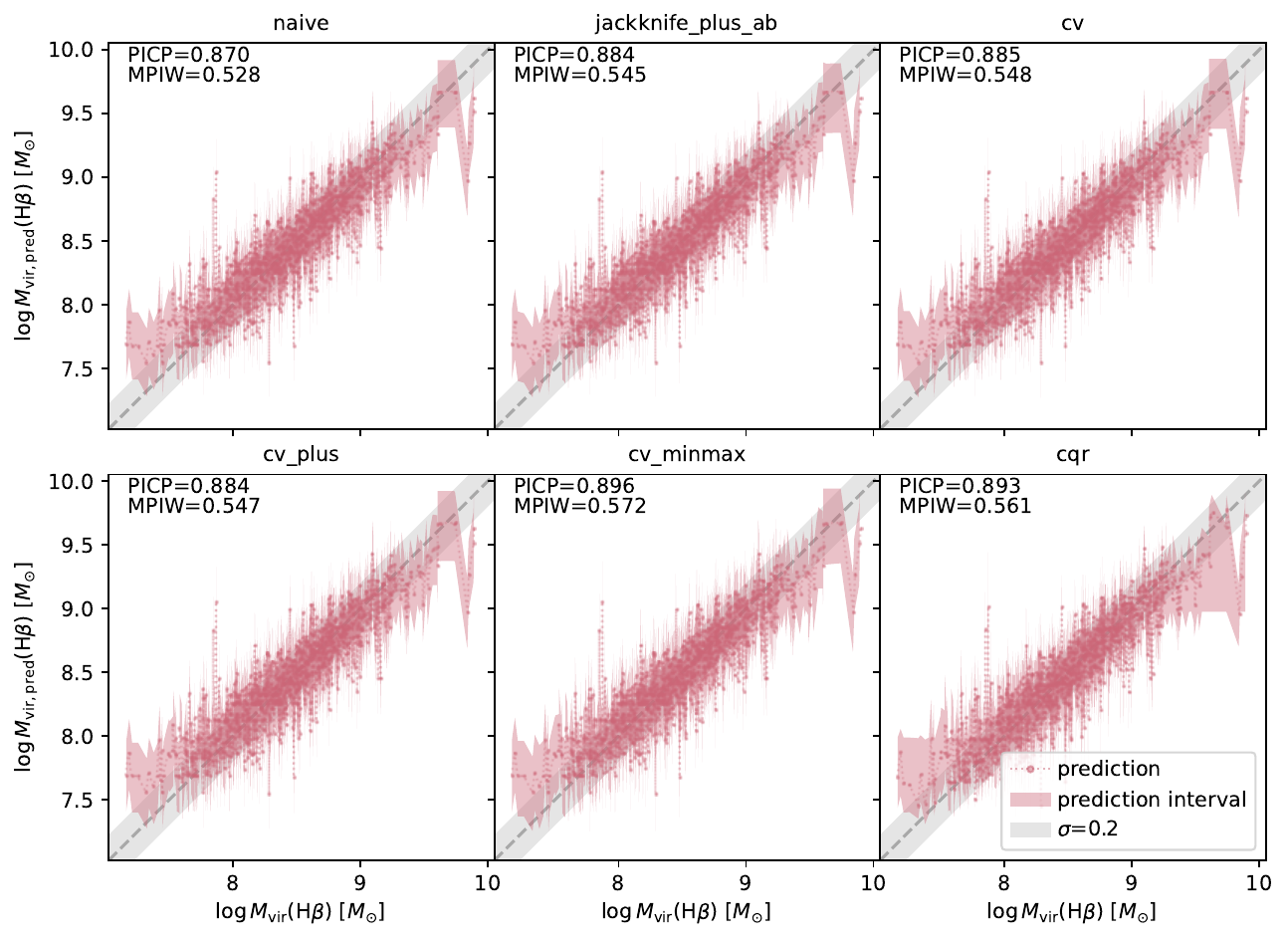}
  \caption{For H$\beta$}
  \label{fig:allpredvstargetuq_hbeta}
\end{subfigure}
\begin{subfigure}[t]{0.83\linewidth}
  \includegraphics[width=\linewidth]{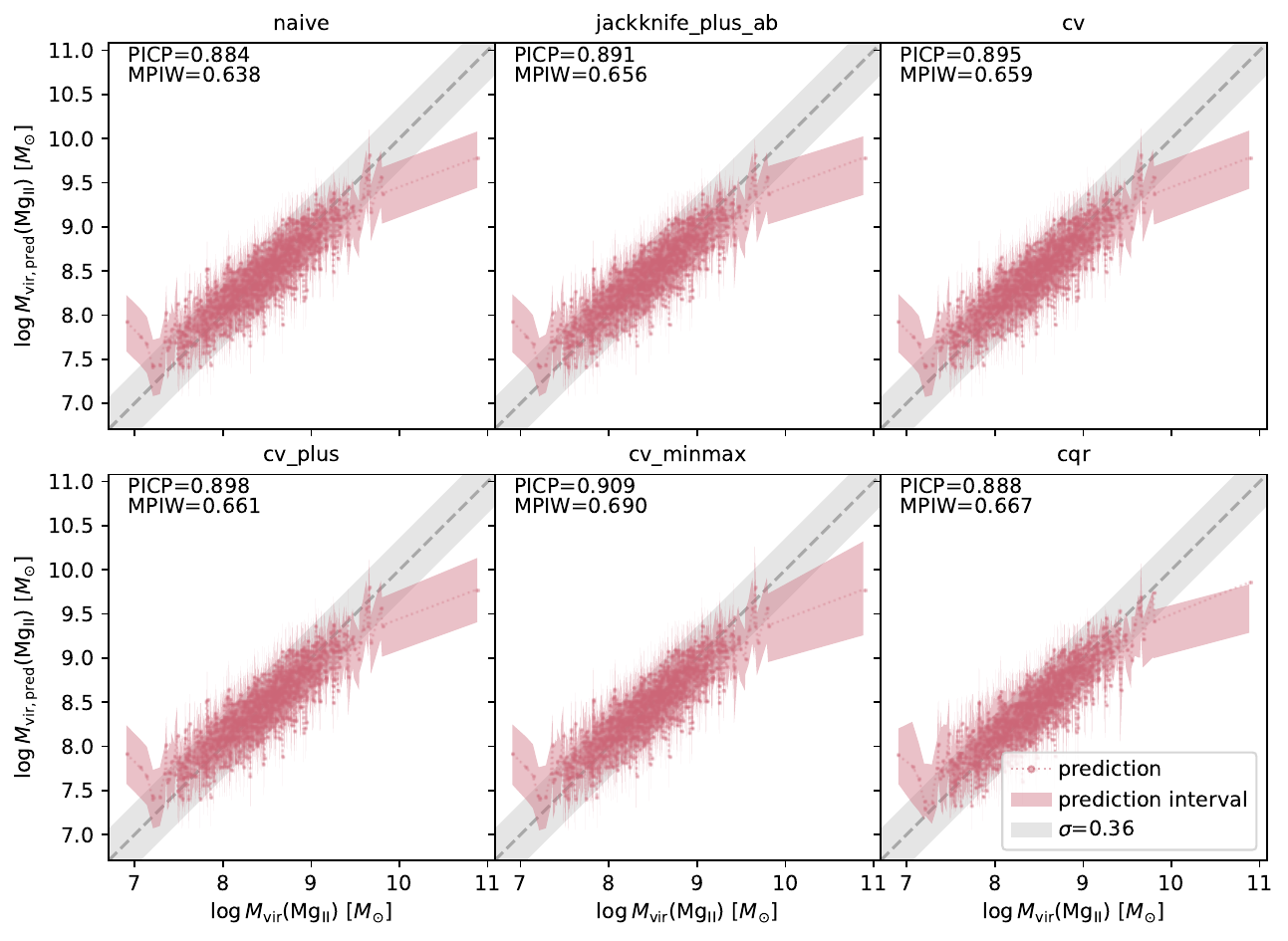}
  \caption{For \ion{Mg}{ii}}
  \label{fig:allpredvstargetuq_mgii}
\end{subfigure}
\caption{Predicted line-based \subtext{M}{vir} comparing different uncertainty quantification methods at coverage of 90\% confidence level.
For reference, the identity line is shown in grey dashed line and the intrinsic scatter from reverberation mapping scaling relation with standard deviation $\sigma$ in shaded gray region.
The predictions are close to those measured from SDSS, implying that the neural network is able to retrieve the \subtext{M}{vir} predictions.}
\label{fig:allpredvstargetuq}
\end{figure*}

A well calibrated uncertainty quantification is valuable to assess the reliability of the black hole mass predictions.
We compare several techniques to estimate the prediction intervals.
The comparison between the predicted mass \subtext{M}{vir,pred} and actual mass \subtext{M}{vir} from SDSS with prediction intervals at 90\% confidence level is presented in \cref{fig:allpredvstargetuq}.
For reference, the shaded gray regions indicate the intrinsic scatter or standard deviation about the scaling relation of 0.2\,dex using H$\beta$ \citep{Bentz+:2013} and 0.36\,dex using \ion{Mg}{ii} line \citep{Homayouni+:2020}.
As previously demonstrated, the neural network is able to retrieve the \subtext{M}{vir} predictions, being comparatively close to those measured from SDSS (\cref{fig:allpredvstargetuq}, identity line in grey dashed line).
All but one of the methods for the \ion{Mg}{ii} line-based \subtext{M}{vir} dataset have PICP lower than the target 90\% confidence level.
Although, this is an indication that they are inadequately calibrated, their PICP remain relatively close to the nominal value.
Overall, at 90\% confidence level, the mean widths of the prediction intervals for all methods are larger than the width of the intrinsic scatter, but still well below some of the reported intrinsic scatter about the $R\text{-}L$ relationship in order of $\gtrsim \pm 0.4\,$dex \citep[e.g.,][]{McLure+Jarvis:2002,Vestergaard+Peterson:2006}, which is $\geq 0.8\,$dex for the width of the scatter.

\begin{figure*}
\centering
\begin{subfigure}[t]{0.49\linewidth}
  \includegraphics[width=\linewidth]{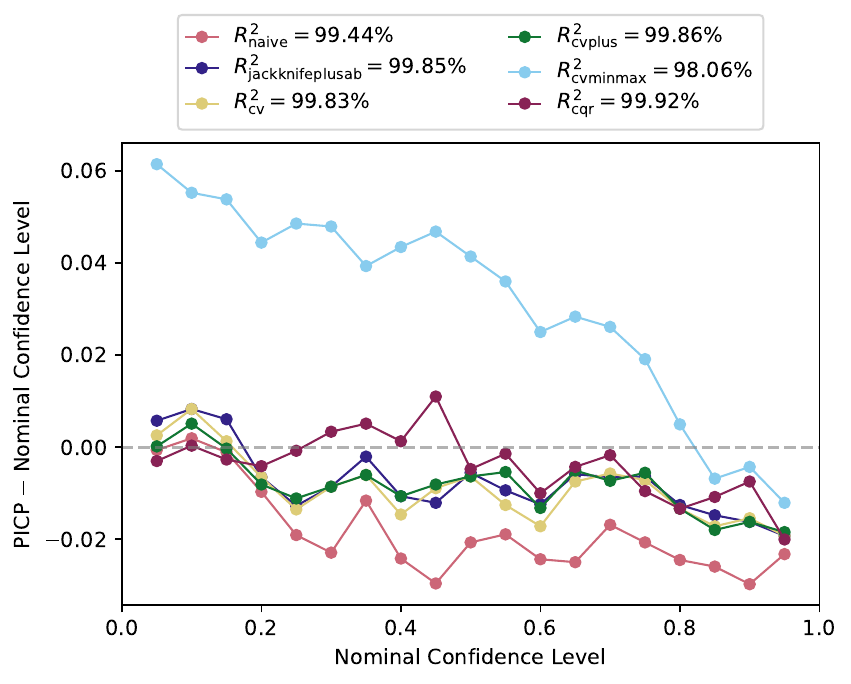}
  \caption{For H$\beta$}
  \label{fig:respicpvsalpha_hbeta}
\end{subfigure}%
\begin{subfigure}[t]{0.49\linewidth}
  \includegraphics[width=\linewidth]{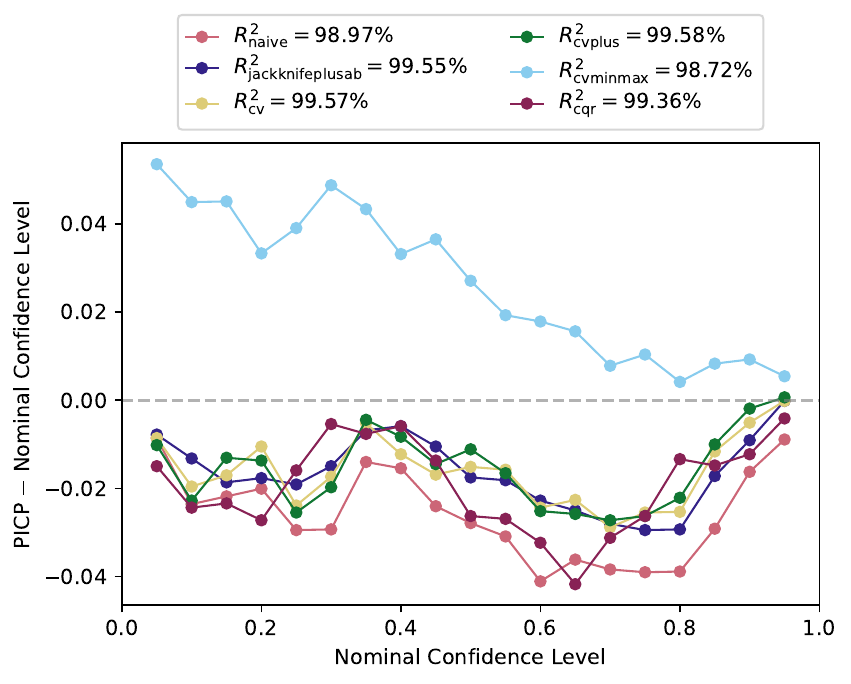}
  \caption{For \ion{Mg}{ii}}
  \label{fig:respicpvsalpha_mgii}
\end{subfigure}
\caption{Difference between prediction interval coverage probability (PICP) and nominal coverage level for various prediction interval methods on line-based \subtext{M}{vir} estimations.
The closer is the line to the horizontal grey dashed line at 0, the better as it implies the PICP achieves nominal coverage, which is also indicated by the higher overall coefficient of determination $R^{2}$.
Above the dashed line is overestimated and vice versa if below.
CV-minmax is the least performing with lowest PICP, followed by naive, and the rest have comparable PICP scores.}
\label{fig:respicpvsalpha}
\end{figure*}

An evaluation of the performance of the prediction intervals over a range of nominal confidence levels using PICP and MPIW is presented next.
As mentioned, it is desirable to have PICP close to the target coverage and small MPIW.
\Cref{fig:respicpvsalpha} displays the difference between the PICP and nominal coverage with respect to the nominal coverage along with the coefficient of determination $R^{2}$ for each method.
For most of the ranges of nominal coverage, the PICP of the naive method is underestimated, while on the other end, the CV-minmax is overestimated.
This is also evident from the lower overall $R^{2}$.
In general, the CV-minmax has the least performing PICP with lowest $R^{2}$, especially when the target confidence level is small.
This is followed by the naive method.
The rest of the methods, including jackknife+ab, CV, CV+, and CQR, have comparable PICP as well as $R^{2}$, particularly towards larger nominal confidence level.

\begin{figure*}
\centering
\begin{subfigure}[t]{0.49\linewidth}
  \includegraphics[width=\linewidth]{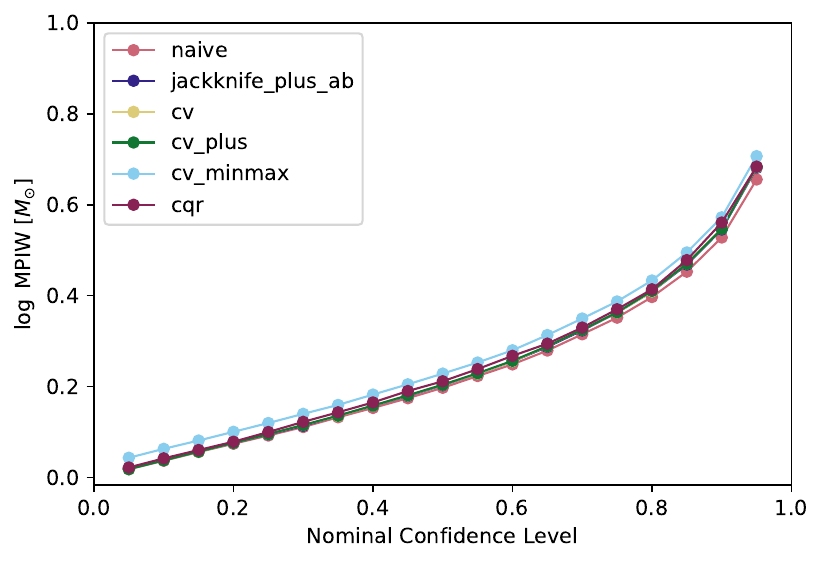}
  \caption{For H$\beta$}
  \label{fig:mpiwvsalpha_hbeta}
\end{subfigure}%
\begin{subfigure}[t]{0.49\linewidth}
  \includegraphics[width=\linewidth]{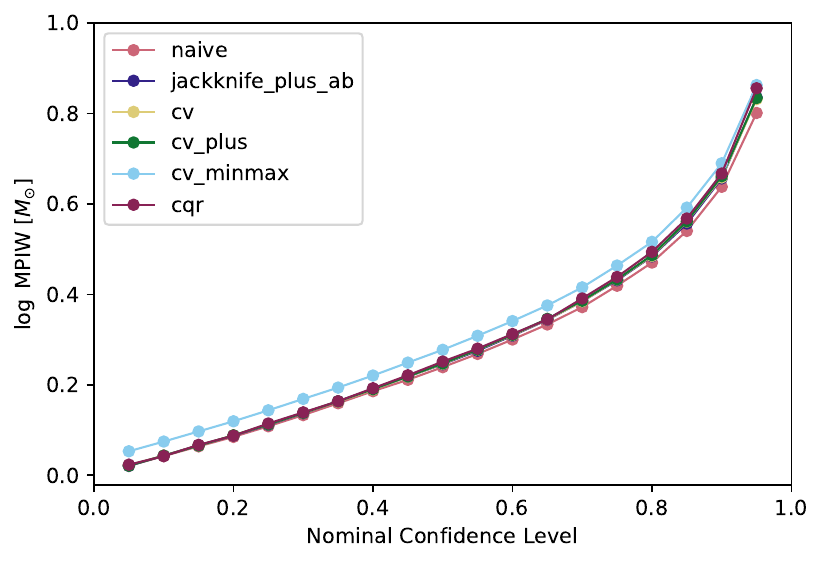}
  \caption{For \ion{Mg}{ii}}
  \label{fig:mpiwvsalpha_mgii}
\end{subfigure}
\caption{Mean prediction interval width (MPIW) for various prediction interval methods on line-based \subtext{M}{vir} estimations.
For all methods, the MPIW scores are relatively close, though the CV-minmax and naive have slightly larger and smaller MPIW, respectively, over a range of nominal coverage.}
\label{fig:mpiwvsalpha}
\end{figure*}

The MPIW scores for a range of nominal coverage is presented in \cref{fig:mpiwvsalpha}.
There is a trade-off of larger width with increasing confidence level, as expected.
The MPIW values for CV-minmax are the largest in all ranges of nominal coverage level, while the MPIW tends to be smaller for the naive method as the nominal coverage is set to be larger.
Other methods have similar MPIW.

\begin{figure}
\centering
\includegraphics[width=\linewidth]{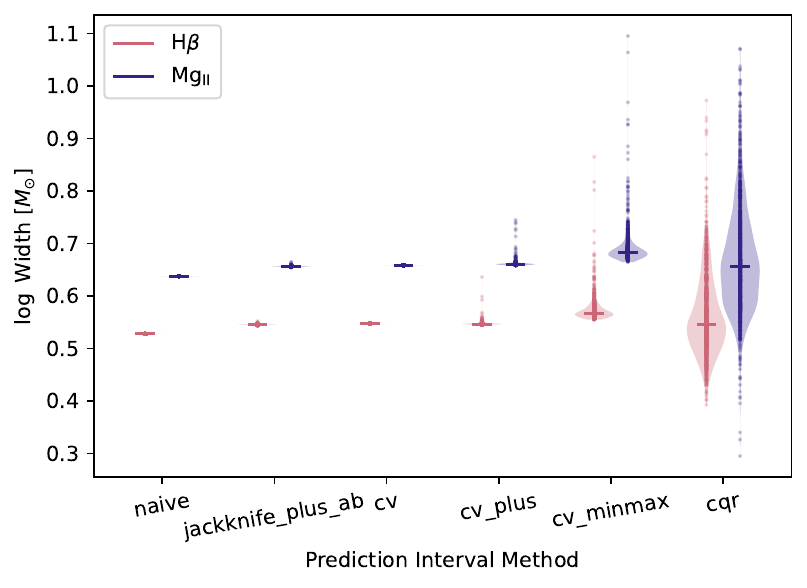}
\caption{Variations in the prediction interval widths for various prediction interval methods at coverage of 90\% confidence level.
The distribution is shown in shaded region with median indicated by the horizontal line and the dots along the distribution are the data points with higher opacity for denser region.
No or negligible variations in the width for naive, jackknife+ab, CV, and CV+.
CV-minmax and CQR show substantial variable widths, though the widths from CV-minmax are always at least wider than those predicted from constant width prediction interval methods.
CQR is more adaptive and can yield narrower widths.}
\label{fig:violinwidth}
\end{figure}

For the remaining of the analysis, the results are for 90\% confidence level, unless otherwise stated.
\Cref{fig:violinwidth} compares the degree of variations in the prediction interval widths for the different uncertainty quantification methods.
The naive, CV, and jackknife+ab methods produce constant or negligible changes in the widths of the prediction intervals.
The prediction bounds from CV+ are also mainly constant except for a few minorities.
The two methods that exhibit variable widths are the CV-minmax and CQR.
However, it can be seen that CV-minmax will generate at the very least larger widths compared to the widths from the constant prediction interval methods as baseline.
Using CQR, it shows greater variability and is able to yield narrower widths under certain circumstances, which will be presented next.
When comparing the scale of the H$\beta$ and \ion{Mg}{ii} \subtext{M}{vir,pred} prediction widths, those from \ion{Mg}{ii} are wider, which is consistent with it being harder to measure, for instance, due to non-virial component \citep[e.g.,][]{Shen+:2008}.

\begin{figure*}
\centering
\includegraphics[width=0.95\linewidth]{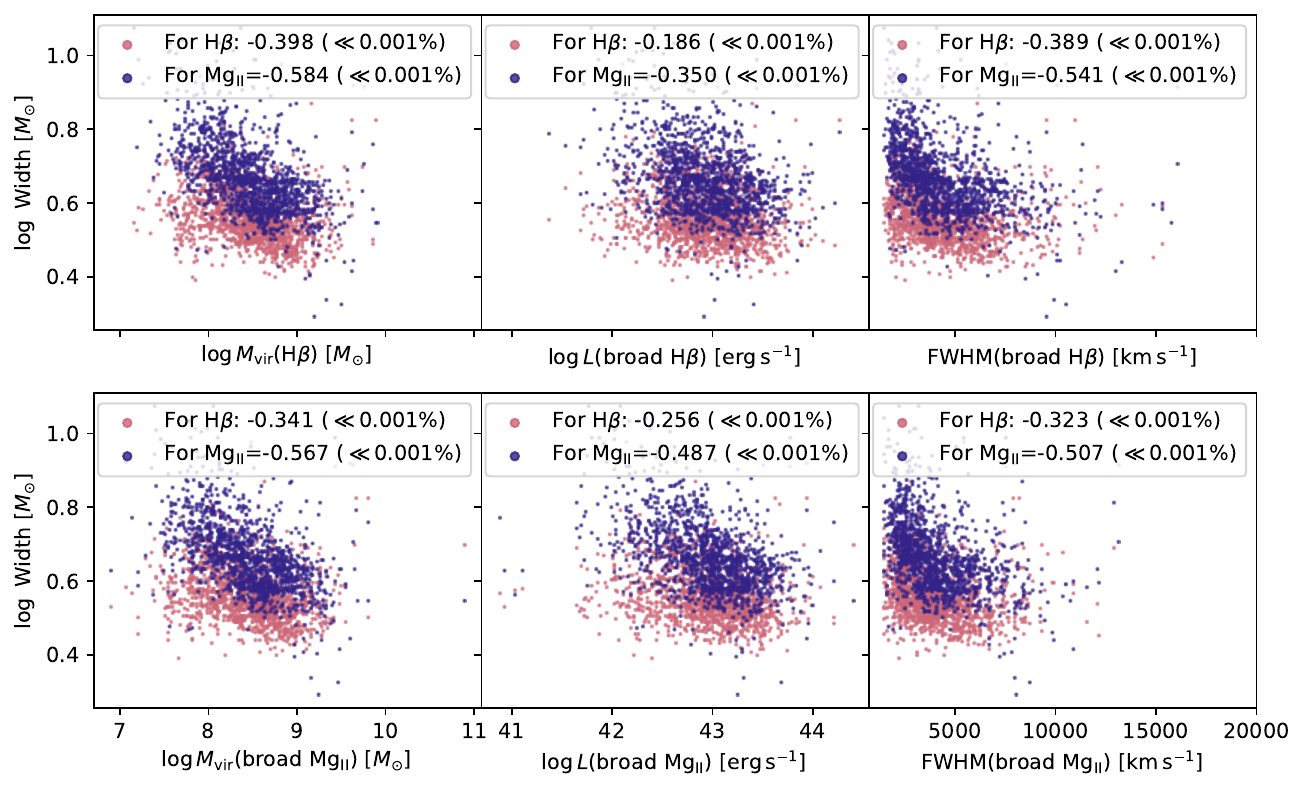}
\caption{Prediction interval widths of H$\beta$ and \ion{Mg}{ii} line-based \subtext{M}{vir,pred} using conformalised quantile regression against selected quasar properties that are related to the black hole mass.
Virial black hole mass in the left panel, line luminosity $L$ in the middle panel, and full width at half maximum (FWHM) of the broad line component in the right panel.
The calculated Spearman's rank correlation coefficient and its corresponding p-value in bracket are shown at the top of each panel.
The higher absolute Spearman's correlation indicates stronger relationship with $|\pm 1|$ being perfect correlation.
There is an anti-correlation between the width and the black hole mass related quasar properties, with tighter widths as the parameter values increase.}
\label{fig:widthvsparams_cqr}
\end{figure*}

Among the explored uncertainty quantification methods, CQR performs the best; therefore, we focus on CQR and demonstrate its adaptiveness with respect to the properties of the quasars.
\Cref{fig:widthvsparams_cqr} portrays the variations in the prediction interval width for selected quasar properties.
To measure the strength of the correlation, the Spearman's correlation coefficient \citep{Spearman:1904} and its corresponding p-value are also calculated.
It is found that there is a negative correlation (statistically significant at p-value$\ll 0.001\%$) between the prediction interval widths and \subtext{M}{vir}, and subsequently with the black hole mass related quasar properties, including the line luminosity $L$ and the FWHM of the broad component of the H$\beta$ and \ion{Mg}{ii} lines.
Some other quasar properties that are also of significantly correlated with the widths (not shown in the figure) are their respective properties measured using the whole line component.

The two quantities, line luminosity and FWHM, relationship with the prediction interval width is expected as these are incorporated into the virial theorem to estimate the \subtext{M}{vir}.
Between the line luminosity and FWHM, the FWHM is more strongly anti-correlated with the size of the prediction interval width, which is a consequence from the virial theorem.
For more luminous and broader spectral line width quasars, the inferred prediction interval using CQR is able to generate a tighter bound.

\begin{figure}
\centering
\includegraphics[width=0.93\linewidth]{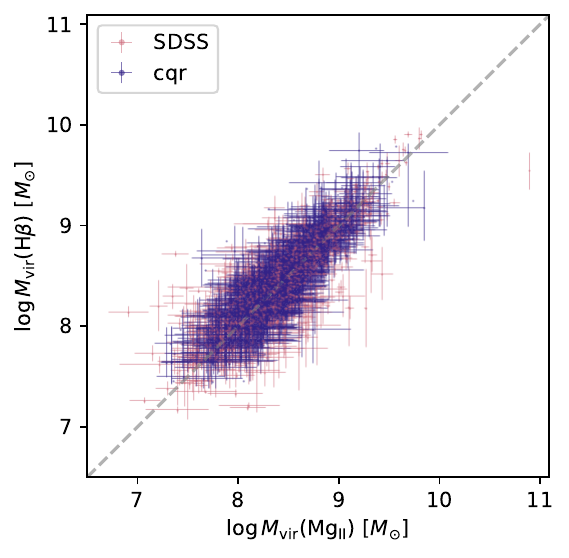}
\caption{Comparison between H$\beta$ and \ion{Mg}{ii} line-based \subtext{M}{vir} from SDSS measurements and conformalised quantile regression (CQR) predictions.
For reference, the identity line is shown in grey dashed line.
The error bars for the SDSS measurements are from measurement errors, while for CQR are from the prediction intervals. 
To avoid clutter, errors are plotted for every third point only.
In both cases, the \subtext{M}{vir} values and error bars are comparable, implying the retrieved values from CQR are in agreement with those from SDSS.}
\label{fig:mvirhbvsmgii_cqr}
\end{figure}

\begin{figure}
\centering
\includegraphics[width=0.93\linewidth]{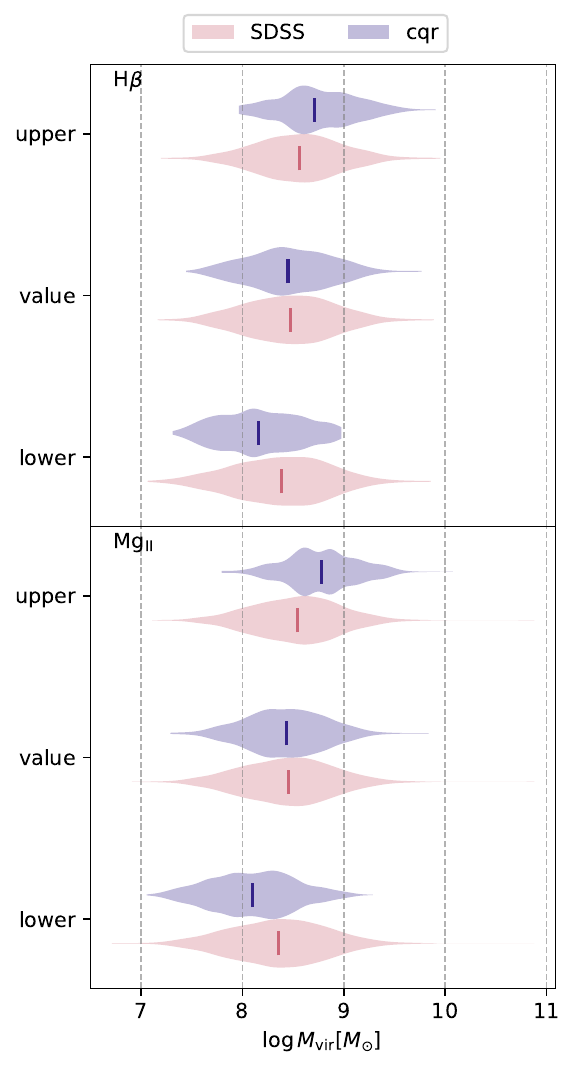}
\caption{Distributions of the line-based \subtext{M}{vir} and their corresponding lower and upper bounds from SDSS measurements and conformalised quantile regression (CQR).
The vertical lines show the median.
The lower and upper bounds for the SDSS measurements are from measurement errors, while for CQR are from prediction intervals.
In both cases, the \subtext{M}{vir} values and error bars are within the same scale, demonstrating that the CQR is able to retrieve the \subtext{M}{vir} comparable to those from SDSS.}
\label{fig:violinpi_cqr}
\end{figure}

We then compare the \subtext{M}{vir,pred} based on H$\beta$ and \ion{Mg}{ii}, as well as their associated prediction intervals using CQR in \cref{fig:mvirhbvsmgii_cqr}.
Comparing multiple emission lines are recommended to get a better constraint of the \subtext{M}{vir} \citep{Vestergaard+:2011}.
Similar analysis is commonly carried out using single-epoch \subtext{M}{vir} calibrated with empirical scaling relation from reverberation mapping \citep[e.g.,][]{McLure+Jarvis:2002,Shen+:2008,Shen+:2011}.
As expected, the H$\beta$ and \ion{Mg}{ii} line-based \subtext{M}{vir} are tightly correlated, albeit the large scatter.
This is not surprising, considering that the amount of scatter from the SDSS measured \subtext{M}{vir} is even larger, as illustrated in \cref{fig:violinpi_cqr}.
Since the errors from SDSS measurements only account for the propagated measurement errors, the median lower and upper intervals are smaller in comparison to the prediction intervals from CQR, as expected.
The retrieved H$\beta$ and \ion{Mg}{ii} based \subtext{M}{vir,pred} along with the prediction intervals using CQR are comparable to those measured from SDSS.

\section{Discussions}\label{sec:discussions}
\subsection{Black hole mass predictions and uncertainties}

In the past decade, artificial intelligence and machine learning have witnessed increasing growth and gained popularity within the astronomy community to solve big data challenges \citep{Baron:2019,Fluke+Jacobs:2020,Huertas-Company+Lanusse:2023}.
Not surprisingly, recently a number of papers have employed machine learning to predict the masses of the black hole in AGN \citep{Chainakun+:2022,He+:2022,Eilers+:2022,Lin+:2023}.
In those studies, they mainly focused on retrieving the predictions of the true black hole mass, whereby the performance in terms of prediction error is usually assessed using MAE, MSE, or RMSE.
Yet, this only evaluates the ability of the machine learning model to recover the true value, but not the reliability of the predictions.

Uncertainty quantification of the black hole mass predictions is vital, especially since the single epoch \subtext{M}{BH} estimates already suffer from a wide range of intrinsic scatter \citep[e.g.,][]{Kelly+Bechtold:2007,Shen+:2008,Shen+Kelly:2010}.
In fact, the uncertainty can extend more than 0.5\,dex for individual AGN \citep[e.g.][]{Vestergaard+Peterson:2006} and is dependent on which emission line is used to probe the \subtext{M}{BH} \citep[e.g.,][]{Peterson+Wandel:1999}.
At the same time, there are also uncertainties from the adopted machine learning pipelines, which introduce further uncertainties into the \subtext{M}{BH} estimation.
Without properly accounting for the uncertainties in the predicted \subtext{M}{BH}, the recovered value will be more biased than it already is.
Therefore it is more desirable to quantify the uncertainties of the black hole masses for each individual AGN rather than for the general AGN population.
Specifically, variable or adaptive widths prediction interval should be considered, as addressed in this study.
Subsequently, one can then attain the prediction interval and conduct analysis similar to those done in reverberation mapping studies (or other black hole mass estimation techniques).

\subsection{Proposed adaptive uncertainty quantification}

We recommend the need to not only assess the performance of the predictions of the black hole mass from the machine learning model, but also quantify the uncertainties for the prediction intervals.
We present an uncertainty quantification method to generate adaptive prediction intervals for the black hole mass estimation using CQR introduced by \citet{Romano+:2019}.
In \cref{sec:results}, we have shown that CQR is more informative of the model's uncertainty compared to other investigated uncertainty quantification methods.
Therefore, we propose that a variable width prediction interval method using CQR is better suited for this particular task.

In assessing the performance of the prediction intervals, it can be seen that the CQR outperforms the rest.
Other methods either produce prediction interval widths that are the same or too wide.
The CQR is more adaptive and better reflects the uncertainty of each individual object.
Additionally, we find that the width of the prediction interval is correlated with the black hole mass and its associated properties, particularly the line luminosity and FWHM.
The larger the black hole mass, the tighter the prediction interval widths.
This suggests that given a bright and broad spectral line source, we should be able to predict the black hole mass with more certainty.
We also highlighted that the virial black hole mass predictions and their corresponding prediction uncertainties generated from the combination of the neural network and CQR architecture, are comparable in scale and magnitude as those measured from SDSS using a spectral line fitting algorithm and reverberation mapping scaling relation with errors from measurements.
The dependence of the spectral line fitting algorithm will affect the continuum and spectral emission line width measurements, effectively biasing the black hole mass estimation \citep[e.g.,][]{Shen+:2008}.
In which case, one can then opt to predict the black hole mass and their associated uncertainties using machine learning coupled with CQR as it offers a somewhat agnostic framework to the fitting of the individual spectral emission lines.

The uncertainty quantification methods that we presented in this study can be deployed with any base machine learning algorithm to quantify the uncertainty of the predicted \subtext{M}{BH}.
The code repository at \repolink\ contains \textsc{Python} scripts to get the data (described in \cref{sec:dataset}), run feature extraction using neural networks and run uncertainty quantification for regression (described in \cref{sec:mvirppi}).
These can be used separately or deployed to existing machine learning methods to generate prediction intervals for the black hole mass predictions (refer to \textsc{MAPIE} documentation for more details).
Additionally, the reproducible outputs for the analysis in this work are also provided.
We include a pre-trained model in \textsc{PyTorch} of the feature extraction method from the supervised neural network model that has been trained on the H$\beta$ and \ion{Mg}{ii} line-based \subtext{M}{vir} dataset from SDSS.
Examples of practical usage include evaluating new datasets, fine-tuning existing networks, and employing the pre-trained model in downstream tasks such as classification and anomaly detection based on the quasar properties.
The generated predictions as well as the uncertainty estimates for the different uncertainty quantification methods are included.
Supplementary \textsc{Python} notebooks including tutorials on usage and data analysis are also provided.

\subsection{Further Experimentations and Caveats} \label{ssec:furtherexpcaveats}

It is apparent that the choice of dataset affects the performance of the prediction intervals.
As aforementioned, we have tested with different input spectra, including the full spectra and those with continuum subtracted, though they performed badly.
Therefore, for our input dataset from SDSS, we use the spectral line flux only that have been continuum subtracted.
This means that the input data still depend on the spectral fitting algorithm and procedure, whereby in this case, to fit and subtract the continuum and extract only the regions with the spectral lines.
As we did not visually inspect the spectra, some of them might also have been fitted poorly.
In this case, the derived properties might also be biased.

Another further constraint is that we choose to use spectra that have both H$\beta$ and \ion{Mg}{ii} lines.
As a consequence, the predictions will not perform well in the absence of any of these lines or if other broad emission lines are present.
One obvious experiment is to evaluate on reverberation mapped samples.
For this purpose, we applied our neural network model on the \ion{Mg}{ii} reverberation-mapped SDSS objects from \citet{Homayouni+:2020}.
Using a subset of their sample that contains both H$\beta$ and \ion{Mg}{ii} lines, we found that the model is able to recover the \ion{Mg}{ii}-based \subtext{M}{vir} reasonably well, however, predicting the \ion{Mg}{ii} reverberation mapping black hole measurements yield larger errors (see \cref{apd:predictbhrm}).
This is due to the fact that \subtext{M}{vir} and \subtext{M}{BH} from reverberation mapping are not directly comparable, whereby the former does not account for the unknown $f$ factor that is known to be unique for individual sources \citep{Decarli+:2008,Pancoast+:2014,Yong+:2016,Yong+Webster:2019}; albeit often assumed to be constant \citep{Collin+:2006,Park+:2012,Woo+:2013,Woo+:2015}.

Since we have not performed a rigorous search for the best regressor, further performance improvement on prediction accuracy could be obtained with more computational resources.
Nevertheless, the basic architecture can act as a baseline and is able to obtain an effective feature extraction that leads to a reasonable prediction of the \subtext{M}{vir}.

We have also conducted experiments using unsupervised learning approach on the same dataset.
We employed a vanilla autoencoder model consisting of a layer of 512 neurons for the encoder and decoder with 8 as the latent dimension for feature extraction.
However, it appears that this model is greatly affected by the presence/absence of other strong broad emission lines, in this case the H$\alpha$ line; thus, outputs higher errors compared to those from the supervised learning approach.

It is also important to point out that the coverages of most of the explored uncertainty quantification methods are below the intended nominal coverage, but mainly still rather close to it.
For our purpose, we did not further attempt to achieve the nominal coverage, which possibly can be fixed by increasing the number of calibration samples \citep{Angelopoulos+Bates:2021}.
\citet{Angelopoulos+Bates:2021} has provided an outline of the procedure and also the method to check for the correct coverage.

\subsection{Future prospects and avenues}

We highlight some future investigations that can be carried out.
Though the availability of reverberation mapped objects is currently limited, the single-epoch black hole mass measurements from these are more precise and better constrained than those calibrated from the scaling relation using H$\beta$ line \citep[e.g.,][]{Homayouni+:2020}.
They can then be used to predict the reverberation mapping \subtext{M}{BH}.
Note that there are various systematic errors from the reverberation mapping method that could lead to poor estimates of the black hole mass by a factor of 3 or $\sim 0.5\,$dex \citep{Krolik:2001}.

There are several advantages of using machine learning to perform predictions of the black hole mass.
We demonstrate that the neural network model is capable of retrieving the \subtext{M}{vir} predictions without having to model individual emission lines of the spectrum and derive their line properties.
The estimates are also comparable to those from SDSS measurements.
Since the machine learning approach is general, one can also apply a similar pipeline to predict other properties of quasars, such as the emission line width, and quantify their uncertainties.
The development of a machine learning model that is completely independent on the spectral fitting algorithm and process would be of interest.

Another benefit that has been mentioned in \citet{Eilers+:2022} is that it avoids the use of the empirical scaling relation between the BLR size and luminosity from reverberation mapping to calibrate line-based \subtext{M}{vir}.
With this, the induced bias from the scaling relation can be mitigated or removed.
Subsequently, the inferred \subtext{M}{vir} for high redshift objects using \ion{C}{iv} might also be less biased.
However, it is noteworthy that there are additional complications when using \ion{C}{iv} line such as it is dominated by outflows \citep{Denney:2012,Coatman+:2017,Yong+:2017} instead of virial motion, which is the basis of the \subtext{M}{vir} estimate.
Hence, it is also important to ensure that the dataset used contains reliable measurements with good signal-to-noise ratio \citep{Denney+:2009}.

We envision that the inclusion of uncertainty quantification will provide a useful assessment of the reliability of the black hole mass predictions trained using machine learning.
The CQR as well as other uncertainty quantification methods that we explored in this work can be incorporated in conjunction with many machine learning models.
A further extension to this work is to explore various or novel uncertainty quantification techniques that would improve the coverage and widths, such as in the presence of limited data and data with large measurement errors.
A potential approach is the conformal predictive system for regression \citep{Vovk+:2017,Vovk+:2018}.
Rather than a single interval, the conformal predictive system estimates the cumulative probability distribution.
In this way, it can be used to further access the trustworthiness of the uncertainty based on the difficulty of the estimates.

\section{Summary}\label{sec:summary}

Measuring an accurate black hole mass has been known to be challenging due to the induced bias from the scaling relation that is used to calibrate the virial black hole mass for high redshift sources.
A reliable tool to determine the uncertainty of the virial black hole mass is important to probe the black hole population and evolution.
In this work, we examine various prediction interval methods, including conformalised quantile regression (CQR), to quantify the uncertainties in the H$\beta$ and \ion{Mg}{ii} line-based virial black hole mass estimation.
The code is publicly available at \repolink.
Using quasar spectra from the Sloan Digital Sky Survey, we train the data on a neural network model for feature extraction using supervised learning, which is then provided to a regressor for predictions.

Among the uncertainty quantification methods that we investigated, the CQR generates a more practical and meaningful range of probable intervals compared to other methods such as jackknife+-after-bootstrap, cross-validation and its variations.
The uncertainty interval of every other methods is either fixed or relatively large.
Conversely, the CQR is able to provide variable width prediction intervals and the tightness of the bounds reflects the correlation with the black hole mass as well as its associated properties.
As objects increase in black hole mass, the size of the prediction interval become narrower.
That is, the prediction bound from CQR will be more certain given a luminous object with broad spectral line width.
Additionally, the neural network architecture coupled with CQR framework are able to retrieve the line-based virial black hole masses and their corresponding errors as wellx as those estimated from the Sloan Digital Sky Survey.
The uncertainty quantification method can be deployed to any machine learning algorithm to assess the quality of the black hole mass predictions, and hence, is recommended.

\section*{Acknowledgements}

We thank the anonymous referee for valuable suggestions on the manuscript.

Funding for the Sloan Digital Sky Survey IV has been provided by the Alfred P. Sloan Foundation, the U.S. Department of Energy Office of Science, and the Participating Institutions. SDSS-IV acknowledges support and resources from the Center for High Performance Computing  at the University of Utah. The SDSS website is \url{www.sdss4.org}.

SDSS-IV is managed by the Astrophysical Research Consortium for the Participating Institutions of the SDSS Collaboration including the Brazilian Participation Group, the Carnegie Institution for Science, Carnegie Mellon University, Center for Astrophysics | Harvard \& Smithsonian, the Chilean Participation Group, the French Participation Group, Instituto de Astrof\'isica de Canarias, The Johns Hopkins University, Kavli Institute for the Physics and Mathematics of the Universe (IPMU) / University of Tokyo, the Korean Participation Group, Lawrence Berkeley National Laboratory, Leibniz Institut f\"ur Astrophysik Potsdam (AIP),  Max-Planck-Institut f\"ur Astronomie (MPIA Heidelberg), Max-Planck-Institut f\"ur Astrophysik (MPA Garching), Max-Planck-Institut f\"ur Extraterrestrische Physik (MPE), National Astronomical Observatories of China, New Mexico State University, New York University, University of Notre Dame, Observat\'ario Nacional / MCTI, The Ohio State University, Pennsylvania State University, Shanghai Astronomical Observatory, United Kingdom Participation Group, Universidad Nacional Aut\'onoma de M\'exico, University of Arizona, University of Colorado Boulder, University of Oxford, University of Portsmouth, University of Utah, University of Virginia, University of Washington, University of Wisconsin, Vanderbilt University, and Yale University.

\textit{Software:} \textsc{Astropy} \citep{astropy:2013,astropy:2018,astropy:2022}, \textsc{Jupyter} \citep{jupyter:2016}, \textsc{MAPIE} \citep{Taquet+:2022}, \textsc{Matplotlib} \citep{matplotlib:2007}, \textsc{NumPy} \citep{numpy:2020}, \textsc{pandas} \citep{McKinney:2010,pandas:2020}, \textsc{PyTorch} \citep{pytorch:2019}, \textsc{scikit-learn} \citep{scikit-learn:2011}, \textsc{SciPy} \citep{scipy:2020} \textsc{UMAP} \citep{McInnes+:2018}.

\section*{Data Availability}

The catalogue and spectroscopic data underlying this article are available in Sloan Digital Sky Survey Data Release 16 quasar properties catalogue at \url{http://quasar.astro.illinois.edu/paper_data/DR16Q/}. The code repository used in this work is publicly available at \repolink.

\bibliographystyle{mnras}
\bibliography{References_bhuq}

\appendix
\section{Dataset Sample Selection}\label{apd:sampleselect}

From the 750,414 SDSS DR16Q spectra, we perform quality cuts, as described in \cref{sec:dataset} of the main paper.
The selection criteria and the corresponding number of spectra after each cut are as follow.
Note that the number of drop out with every cut depends on the ordering in which the criterion is performed as there will be spectra that satisfy multiple criteria.
\begin{itemize}
  \item H$\beta$ line flux/flux error $>2$: 140,172
  \item and \ion{Mg}{ii} line flux/flux error $>2$: 133,772
  \item and H$\beta$ logarithm line luminosity ranges 38--48\,erg\,s$^{-1}$: 132,543
  \item and \ion{Mg}{ii} logarithm line luminosity ranges 38--48\,erg\,s$^{-1}$: 132,538
  \item and signal-to-noise ratio per pixel $\geq 10$: 25,283
  \item and H$\beta$ line width available: 25,283
  \item and \ion{Mg}{ii} line width available: 25,283
  \item and H$\beta$ black hole mass available: 14,798
  \item and \ion{Mg}{ii} black hole mass available: 14,777
  \item and H$\beta$ black hole mass error $<0.5$: 14,602
  \item and \ion{Mg}{ii} black hole mass error $<0.5$: 14,314
  \item and H$\beta$ line width error $<2000\,$km\,s$^{-1}$: 14,124
  \item and \ion{Mg}{ii} line width error $<2000\,$km\,s$^{-1}$: 13,952
\end{itemize}
The final data sample contains 13,952 spectra. The SDSS DR16Q data along with the derived catalogue are publicly available at \url{http://quasar.astro.illinois.edu/paper_data/DR16Q/}.

\section{Predicting Black Hole Mass from Reverberation Mapping}\label{apd:predictbhrm}

To examine the performance of the predictions and prediction intervals on reverberation mapped black hole masses, \subtext{M}{RM}, we utilise the reverberation mapped SDSS samples from \citet{Homayouni+:2020} measured using \ion{Mg}{ii} lags.
We cross-match them with the SDSS DR16Q quasar properties catalogue \citep{Wu+Shen:2022} and further restrict those with both H$\beta$ and \ion{Mg}{ii} lines.
This provides a total of 14 samples and 7 among them are gold samples with most credible \ion{Mg}{ii} lags of $\leq 10$\% individual false positive rate.

\begin{figure}
\centering
\includegraphics[width=\linewidth]{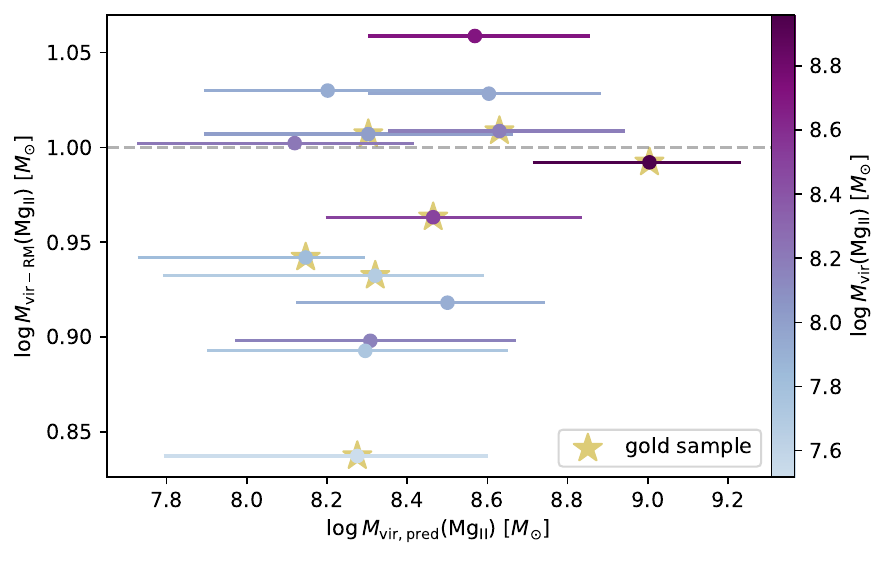}
\caption{Black hole mass for \ion{Mg}{ii} predictions and measurements comparing those from virial estimates, \subtext{M}{vir}, and reverberation mapping \subtext{M}{RM}, coloured by \ion{Mg}{ii}-based \subtext{M}{vir} with darker gradient being larger mass.
The corresponding prediction intervals are from conformalised quantile regression.
The gold sample with most credible \ion{Mg}{ii} lags of $\leq 10$\% individual false positive rate in star.
The difference between \subtext{M}{vir} and \subtext{M}{RM} can be up to $\sim 1.5$\,dex.}
\label{fig:diffmbhvirrmvspred_cqr}
\end{figure}

\Cref{fig:diffmbhvirrmvspred_cqr} shows the comparison between the \subtext{M}{vir} and \subtext{M}{RM} along with the predictions from the supervised neural network and prediction intervals from conformalised quantile regression, as outlined in \cref{sec:mvirppi} of the main paper.
Half of the samples are gold samples (\cref{fig:diffmbhvirrmvspred_cqr}, gold star).
The majority of them have \subtext{M}{vir} close to \subtext{M}{RM} within 0.5\,dex, except one that also shows the largest discrepancy of $\sim 1.5$\,dex.
As mentioned, this inconsistency arise due to the fundamental difference between the \subtext{M}{vir} and \subtext{M}{RM}.
A further suggestion is to train using \subtext{M}{RM} samples in order to predict the same quantity.

\bsp	
\label{lastpage}
\end{document}